\documentclass[letterpaper, 10 pt, conference]{ieeeconf} 

\usepackage{hyperref}
\usepackage{cite}
\usepackage[font=footnotesize]{subfig}
\usepackage{multirow}
\usepackage{ifpdf}

\usepackage{epstopdf}
\usepackage[cmex10]{amsmath}
\usepackage{algorithmic}
\usepackage{array}
\usepackage{mdwmath}
\usepackage{mdwtab}
\usepackage{eqparbox}
\usepackage{verbatim}
\usepackage{booktabs}
\usepackage{multirow}
\usepackage{graphicx}
\usepackage{algorithm}
\usepackage{algorithmic}
\usepackage{array} 
\usepackage[font=footnotesize]{subfig}
\usepackage{fixltx2e}
\usepackage{stfloats}

\usepackage{amsmath,bm}
\usepackage{amssymb}
\usepackage{color}
\usepackage{subfig}

\usepackage{multirow}
\usepackage{float}
\usepackage{caption}
\usepackage[normalem]{ulem}
\usepackage{comment}
\definecolor{orange}{rgb}{1,0.5,0}

\begin{document}

\title{\LARGE \bf  TIMS: A Tactile Internet-Based Micromanipulation System with Haptic Guidance for Surgical Training}
 \author{Jialin Lin,  Xiaoqing Guo, Wen Fan, Wei Li,   Yuanyi Wang, Jiaming Liang, Jindong Liu, Weiru Liu,  Lei Wei, \\ Dandan Zhang\\
\thanks{
J. Lin, X. Guo, W. Fan, W. Liu, and D. Zhang, are with Engineering Mathematics, University of Bristol, affiliated with the Bristol Robotics Lab, United Kingdom. 
W. Li is with the Hamlyn Centre for Robotic Surgery, Imperial College London, United Kingdom. D. Zhang and J. Liu are honorary researchers at Imperial College London. Y. Wang, J. Liang, and L. Wei are with Tencent Robotics X.  Corresponding Author: Dandan Zhang (ye21623@bristol.ac.uk, d.zhang17@imperial.ac.uk). }
}

\maketitle

\begin{abstract}



Microsurgery involves the dexterous manipulation of delicate tissue or fragile structures, such as small blood vessels and nerves, under a microscope. To address the limitations of imprecise manipulation of human hands, robotic systems have been developed to assist surgeons in performing complex microsurgical tasks with greater precision and safety. However, the steep learning curve for robot-assisted microsurgery (RAMS) and the shortage of well-trained surgeons pose significant challenges to the widespread adoption of RAMS. Therefore, the development of a versatile training system for RAMS is necessary, which can bring tangible benefits to both surgeons and patients.

In this paper, we present a Tactile Internet-Based Micromanipulation System (TIMS) based on a ROS-Django web-based architecture for microsurgical training. This system can provide tactile feedback to operators via a wearable tactile display (WTD), while real-time data is transmitted through the internet via a ROS-Django framework. In addition, TIMS integrates haptic guidance to `guide' the trainees to follow a desired trajectory provided by expert surgeons. Learning from demonstration based on Gaussian Process Regression (GPR)  was used to generate the desired trajectory. 
We conducted user studies to verify the effectiveness of our proposed TIMS, comparing users' performance with and without tactile feedback and/or haptic guidance.
For more details of this project, please view our website: \url{https://sites.google.com/view/viewtims/home}. 
\end{abstract}


\section{Introduction}

Microsurgery is a type of surgery that involves manipulating delicate tissues under an operating microscope. It is a specialized surgery that has high requirements for surgeons' skills and attention \cite{zhang2022teleoperation}.
Robot-assisted microsurgery (RAMS) is an emerging field of surgical innovation that uses microsurgical robots to improve precision and dexterity for microsurgery, leading to better clinical outcomes \cite{zhang2019handheld}. More specifically, RAMS aims to overcome the physical limitations of human operators in microsurgery \cite{payne2021shared}, which leads to the reduction of  hand tremors and the improvement of patient outcomes \cite{gao2021progress}.

\begin{figure}[htbp]
    \centering
     \captionsetup{font=footnotesize,labelsep=period}
    \includegraphics[width=\linewidth,scale=1.00]{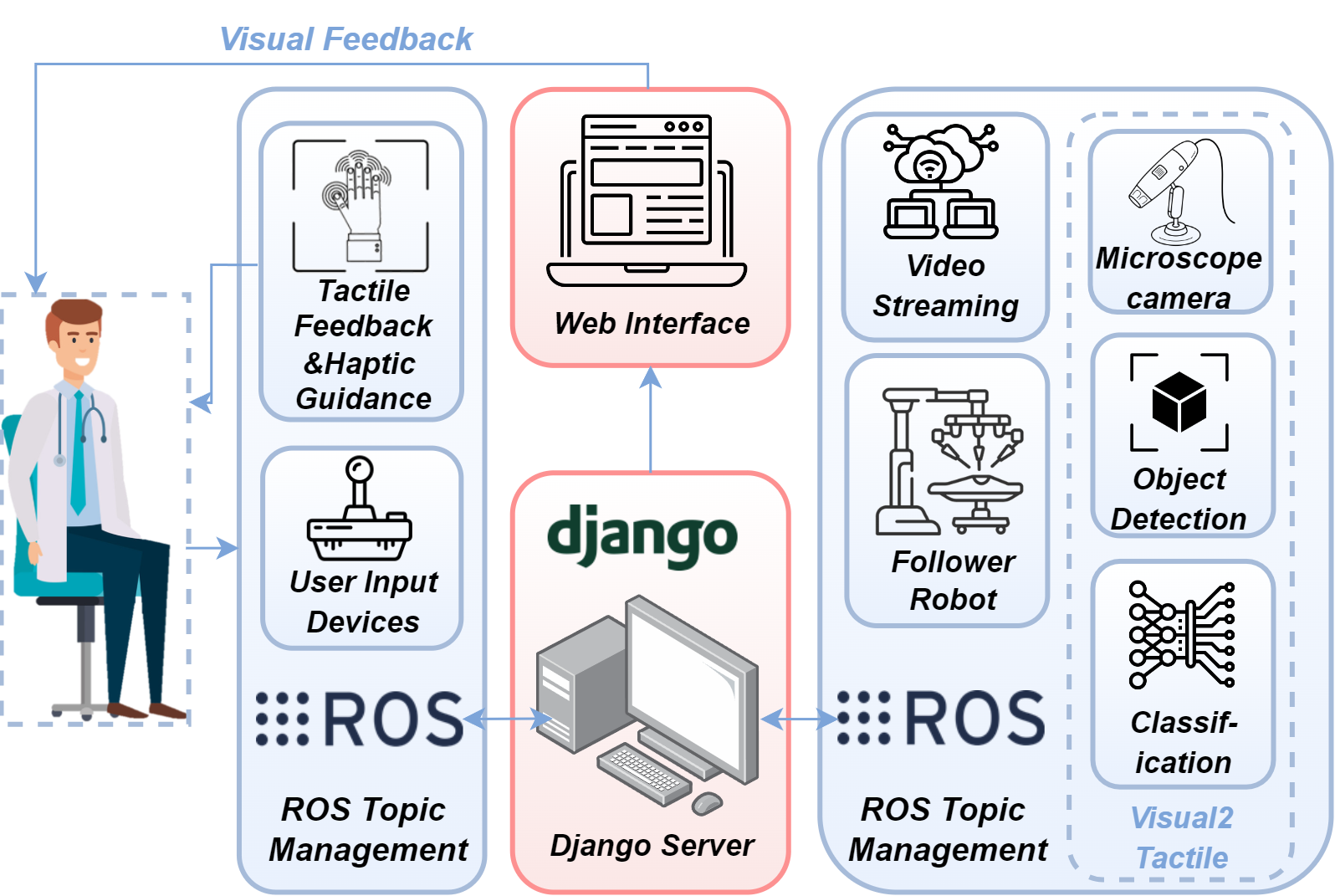}
    \caption{TIMS system overview, including the illustration of control interface, feedback system, ROS-Django framework, data transmission.}
     \vspace{-0.4cm}
    \label{fig:SystemDesign}
\end{figure}



To perform RAMS with high proficiency, microsurgical trainees need to go through extensive practice before applying the learned skills to patients in clinical settings.   However, the lack of available training platforms and the high cost of setting up microsurgical platforms make it challenging to train new surgeons within a short period of time \cite{lallas2012robotic}. Therefore, it's significant to develop a versatile and efficient surgical training platform that can help surgeons improve their proficiency in using microsurgical platforms \cite{puliatti2020training}.
To this end, there is an increasing demand for microsurgical skill training outside the clinical setting to reduce the occupation of the clinical resource \cite{malik2017acquisition}.  

For microsurgical training, simulation offers a solution to  enable trainees to acquire basic microsurgical skills in a home-based environment. However, most of the existing simulators for RAMS  are of low fidelity \cite{malik2017acquisition}. Though simulators can be used to provide initial surgical skill training, we aim to investigate physical training systems to provide a high-fidelity learning environment for trainees. Well-equipped microsurgical laboratories are not so widespread as their setups are expensive \cite{choque2018virtual}. With this regard, we aim to develop an accessible cost-effective training system, allowing trainees to practice and take ownership of their technical skill development. We envision that with the microsurgical training system, novice surgeons can master basic micromanipulation skills and get familiar with microsurgical sub-tasks within a relatively short time \cite{ilie2008training}.
A low-cost microsurgical robot research platform (MRRP) has been developed for microsurgical skill training and assessment \cite{zhang2020microsurgical}. However, MRRP does not include haptic guidance functions, and cannot provide tactile feedback to users. It also lacks remote training capabilities, which means that expert surgeons cannot supervise surgical trainees who are located in different geographic locations. To address these limitations, we aim to develop a new RAMS training platform that allows novice physicians to receive remote training over the internet.


Most existing RAMS training platforms rely solely on visual feedback \cite{zhang2022human}.  The  essential objectives in human-in-the-loop teleoperation systems are the enhancement of user awareness about the robot's state and the optimization of teleoperation fidelity. To accomplish these goals, the provision of haptic feedback, encompassing both force and tactile feedback, becomes indispensable. \cite{giri2021application}.  For safety considerations in RAMS, virtual fixtures can be constructed to protect fragile tissues from damage caused by extra-exerted force \cite{chen2020supervised}. With virtual fixtures, force feedback can be provided to surgeons once the microsurgical tools exceed the safety boundary.  The concept of the `Tactile Internet' has been proposed to enable remote operators to interact with target objects using tactile feedback \cite{fettweis2014tactile}. Tactile Internet is particularly significant for telesurgery, where the sense of touch is critical for surgeons.  Although active constraints or virtual fixtures can help ensure safety when surgeons lose their sense of touch during telemanipulation tasks, direct tactile feedback is still crucial to surgeons when performing surgery to prevent tissue injury or suture breakage.  Therefore, tactile internet should be integrated into the microsurgical training platform \cite{Piriyanont2013Design}.




In addition, haptic guidance is crucial for surgical training since it can enable operators to manipulate a microsurgical robot along a predetermined trajectory \cite{power2015cooperative}. In situations where the operator deviates from the predetermined trajectory, the haptic device  can generate a force to guide the operator's hand back onto the correct path. This is similar to a teacher guiding a student's hand as they learn to write. The incorporation of haptic guidance in the system for surgical training can enable trainees to quickly learn how to perform microsurgery. 
Therefore, we aim to incorporate haptic guidance into our proposed system to improve the effectiveness and efficiency of surgical training.




In summary, we propose the Tactile Internet-Based Micromanipulation System (TIMS) with Haptic Guidance for Surgical Training in this paper. 
The proposed system can reduce surgeons' workload for perceiving tool-tissue interaction during microsurgical operation and help minimize potential damage to delicate tissues during RAMS.
 The \textbf{main contributions} of this paper are as follows:
\begin{itemize}
    \item We developed a relatively low-cost micromanipulation system, which can be used for high-fidelity microsurgical training.
    \item We developed the infrastructure of Tactile Internet for RAMS, which can provide real-time tactile feedback to surgical trainees. 
    \item We constructed a haptic guidance framework for microsurgical training using learning from demonstration techniques.
\end{itemize}
 TIMS represents the first step toward Microsurgical Skills Self-Training Laboratory with cost-effective microsurgical instruments \cite{chung2017affordable}.
To the best of our knowledge, this is the first affordable high-fidelity microsurgical training system that integrates haptic guidance and tactile internet technology.

\section{Methodology}
\subsection{System Overview}
As shown in Fig. \ref{fig:SystemDesign}, the proposed TIMS is developed based on a Python-based ROS-Django framework, which combines the flexibility of ROS in managing multiple hardware resources. This system is designed to improve surgical trainees' performance in surgical training by integrating tactile feedback and haptic guidance. A bilateral teleoperation system is therefore constructed, with a forward loop for microsurgical robot control and a backward loop for  providing haptic and visual feedback to trainees.
The proposed TIMS has several key features that make it user-friendly and effective for surgical trainees:

\textbf{Tactile Internet:} Most of the teleoperation system provides operators with only visual feedback \cite{zhang2019design}. In our proposed system, haptic rendering is seamlessly integrated into the system, which combines kinaesthetic rendering (force feedback) through a commercial device with tactile rendering through an in-house wearable tactile display (WTD). 
Moreover, the Django framework enables both robot control messages and real-time microscopic video to be transmitted to the web page, allowing the trainee to remotely perform microsurgical training with vision and haptic feedback simultaneously. Thanks to the user-friendly web-based architecture, surgeons can supervise trainees remotely.


\textbf{Haptic Guidance:} 
The role of haptic guidance is to guide trainees back to the desired path when they deviate from the correct trajectory through  the haptic guidance force. The haptic guidance force is generated through the comparison of the real-time operation trajectory and the desired trajectory generated by expert surgeons through learning from demonstration. Moreover, the haptic guidance can also provide the operators with force feedback if the position of the surgical tools exceeds the safety boundary.


\textbf{Automatic Skill Analysis:}  The automatic surgical skill analysis  allows trainees to receive real-time feedback on their performance during training sessions \cite{zhang2020automatic}.  The backend database attached to the ROS-Django framework can automatically record all operation data performed by the trainee. 
This data can be used for post-surgery operational analysis, and the results will be automatically displayed on a user-friendly web-based interface, allowing trainees to evaluate their learning outcomes.

\subsection{Hardware Description}

The microsurgical training system is constructed based on bilateral teleoperation, which is built upon the leader-follower control mode.
\begin{figure*}
    \centering
    \captionsetup{font=footnotesize,labelsep=period}
    \includegraphics[width = 0.95\hsize]{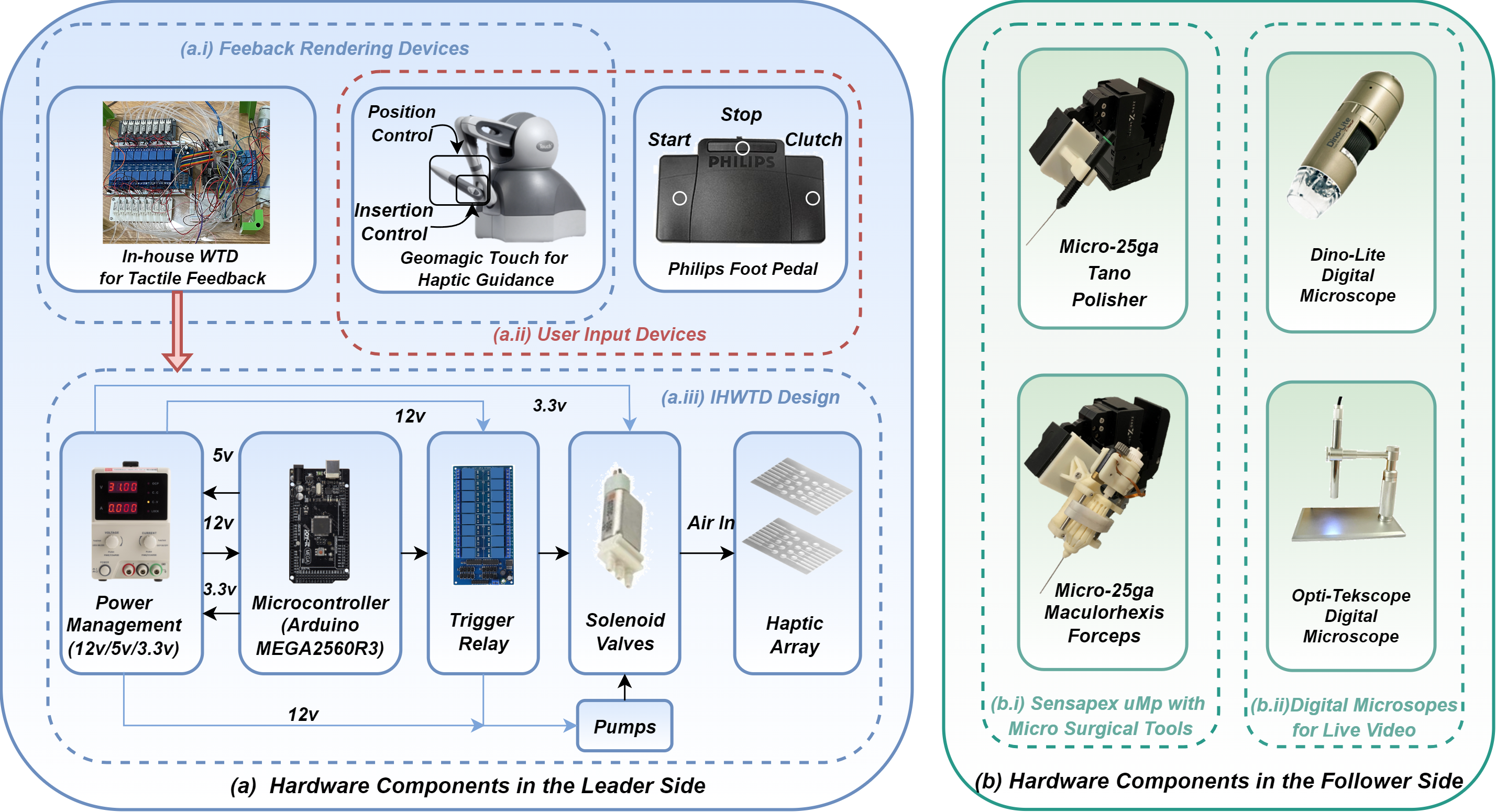}
    \caption{Illustration of the hardware components in the TIMS system. (a) shows the hardware components on the leader side, including  the commercial motion capture device (Geomagic Touch), the wearable touch display and  the foot pedal. (b) shows the microsurgical robot, digital microscopes, and different microsurgical tools used in the experiments.}
  \vspace{-0.3cm}
    \label{fig:hardware}
\end{figure*}

\textbf{Hardware Components in the Leader Side:}


The hardware system of the TIMS leader side comprises three major components, including i) a motion capture device with kinaesthetic rendering capability, ii) a foot pedal for system logistics control, and iii) an in-house wearable touch display (WTD) (an essential component of the tactile internet).

Geomagic Touch (3D System), a commercial haptic device, is used for motion capture and kinaesthetic rendering in the TIMS system.   This device accurately tracks and replicates the surgeon's hand movements and provides kinaesthetic feedback to the surgeon if the surgical tool falls into forbidden regions. The foot pedal, one of the most common user interfaces for robotic surgery, is used to manage system logistics, such as starting or stopping the training system and enabling teleoperation for robot control.

The in-house WTD is constructed using a pneumatically actuated tactile actuator array.   It consists of two film layers, which are made from a pliable material called Thermoplastic Polyurethane (TPU).  Each pneumatically actuated tactile actuator array contains 16 inflatable actuators arranged in a 2D pattern. These actuators are essentially cylindrical air pockets, measuring 3 mm in diameter. 
This display has rectangular dimensions of approximately 30 mm by 20 mm, which corresponds to the size of human fingertips. By injecting air into an actuator's chamber, the user can feel a tactile sensation generated by the inflatable actuators. The components used for constructing this wearable tactile display can be viewed in Fig. \ref{fig:hardware} (a.iii).



\textbf{Hardware Components in the Follower Side:}
A micromanipulator (Senasapex uMp) is used as the high-precision position stage to mount different microsurgical tools for different types of microsurgical training.   As shown in Fig. \ref{fig:hardware} (b.i),  a motorized 25ga Maculorhexis Forceps (Katalyst Surgical, LLC, USA) and a 25ga Tano Polisher (Katalyst Surgical, LLC, USA) can be mounted on the stage to conduct high-precision micromanipulation for different microsurgical sub-tasks, such as membrane peeling and needle insertion, etc. The microsurgical robot is mounted on an anti-vibration optical table.
Fig. \ref{fig:hardware} (b.ii) shows the low-cost digital microscopes used in TIMS. The Opti-Tekscope microscope camera is used to capture operation videos from the top-down view, which provides real-time visual feedback to  operators. The Dino-Lite Premier camera is used to capture the side view of the operation scene and its functions will be detailed in Section II.D.




\subsection{ROS-Django Framework}



Robot Operating System (ROS) is a widely used middleware that facilitates peer-to-peer communication among multiple robots. In this system, we used the topic publishing/subscribing feature of the ROS framework to build i) the Geomagic Touch-ROS connection, ii) the Footpedal-ROS connection, iii) the WTD-ROS connection, and iv) the microrobot-ROS connection. ROS continuously publishes and subscribes to topics, thereby organizing and uploading the robot's state information to the Django server.

\begin{figure}[tb]
	\centering
\captionsetup{font=footnotesize,labelsep=period}
	\includegraphics[width = 1\hsize]{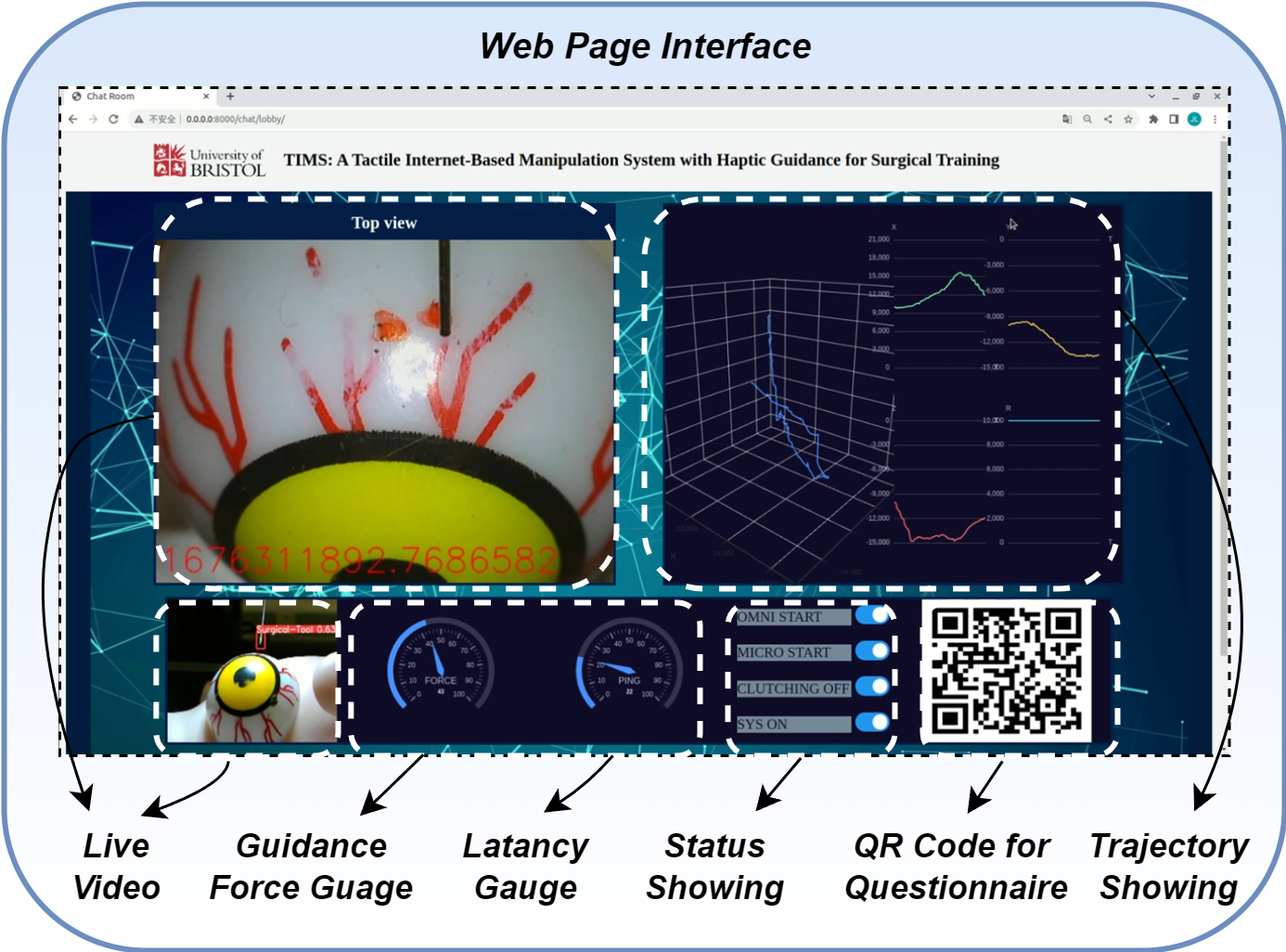}
	\caption{Overview of Web Page Interface constructed based on the ROS-Django Framework.}
  \vspace{-0.4cm}
	\label{fig:web-interface}
\end{figure}

Django is an open-source web framework written in Python.  In the TIMS, Django serves as a `server' responsible for transferring, recording, and displaying robot messages in real time on the website. 
The Django server uses Redis as the database to store all the sensory information and control commands generated during microsurgical operation.
Through the website, the user can observe the robot's state and live microsurgery video, as shown in Fig. \ref{fig:web-interface}.

To communicate with robots, Django establishes a WebSocket connection with ROS.  It can continuously receive JSON strings from ROS and stores the data in Redis by parsing and reading the device ID (Geomagic Touch, Microrobot) in the JSON strings.  In a separate thread, Django asynchronously retrieves the latest message from Redis and sends it to the specific device.  WebSocket is chosen due to its full-duplex communication, which enables the server to forward messages directly to the client. 
In addition, when the Web interface is established, the website can also directly obtain/update messages from the server by connecting to WebSocket. Redis is chosen as the backend database because it is RAM-based, making it faster to read and write than other databases. This property helps reduce latency issues in teleoperation via the internet. As for the video streaming, we use UDP socket to ensure visual feedback with low latency (around 34ms in the local area network).

\subsection{Bilateral Teleoperation}
\subsubsection{Leader-Follower Mapping}







Leader-follower mapping is essential for bilateral teleoperation systems, which defines the data transmission and mathematical formula of mapping the leader robot's motions to the follower robot's end-effector motions. On the leader robot side, ROS acquires real-time messages from  the 3D coordinate of Geomagic Touch's end-effector,  the stylus button status, and the foot pedal status.
These messages are sent to the Django server via WebSocket. The Django Consumer, an abstraction used to implement WebSocket in Django, parses the message, stores it in the database, and then forwards it to the website and the follower robot side. On the follower robot side, ROS publishes these messages to  control the microrobot based on a predefined leader-follower mapping strategy, the mathematical formula of which is as follows:
\begin{equation}
P_{\text{follower}}^i=\alpha\left(P_{\text{leader}}^i-\ P_{\text{leader}}^{i-1}\right)+P_{\text{follower}}^{i-1}
\end{equation}
where $i$ indicates the time step, $P_{\text{follower}}^i$ is the new position of the  microrobot (the follower), $P_{\text{follower}}^{i-1}$ is the previous position of the microrobot. $ P_{\text{leader}}^i$ and $P_{\text{leader}}^{i-1}$ are the new and previous positions of the end-effector of the Geomagic Touch (the leader) respectively. The  $\alpha$ is a motion scaling factor that adjusts the ratio of the position mapping between the leader and follower \cite{zhang2018self}.





\subsubsection{Tactile Feedback}

Tactile feedback will be provided to the operator when the microrobot's end effector touches the simulated tissue during microsurgical training. To provide reliable tactile feedback, the tool-tissue interaction status monitoring is significant.  To circumvent the need for attaching expensive micro-sensors to the microsurgical tooltip, we propose an alternative solution, i.e. developing a deep learning-based force estimation approach based on vision data \cite{wang2021real}.
More specifically, a digital microscope is used to capture the side view of the phantom for surgical training. The captured images are subsequently transformed into tactile information using a deep learning-based object (tooltip) detection and classification approach, which serves to monitor whether the tooltip comes into contact with delicate tissues or not. 


Yolov5 \cite{glenn_jocher_2021_5563715} is employed in this system for tooltip tracking.  2,649 images were collected and labeled manually, including 80\% used as the training set, and 10\% each as the validation and test sets. The model achieved its best performance with a precision of 0.901, recall of 0.982, and mAP50 of 0.893 on the test set. 
The frame is cropped using the output bounding box from the Yolov5 model. This ensures that the surgical tooltip becomes the dominant element in the resulting image rather than the image background, thereby significantly reducing the difficulty of image classification in the following step.
We then employ a four-layer convolutional neural network (CNN) for image classification, attaining an impressive accuracy of 99.8\% during real-time  object (tooltip) detection tasks. This model outputs a boolean value to determine whether the microsurgical tool touches the simulated tissue or not. The ROS-Django framework transmits this boolean value to the in-house WTD via the internet, enabling the rendering of tactile sensation on the operator's fingertip. When the `touch' signal is received, the pump of the WTD starts to inflate all the air pockets (inflatable actuators) to provide tactile feedback. When the signal is `not touch', the pump stops working and the air in the device is released.


 \subsubsection{Haptic guidance}
 
\begin{figure*}[tb]
	\centering
\captionsetup{font=footnotesize,labelsep=period}
	\includegraphics[width = 1\hsize]{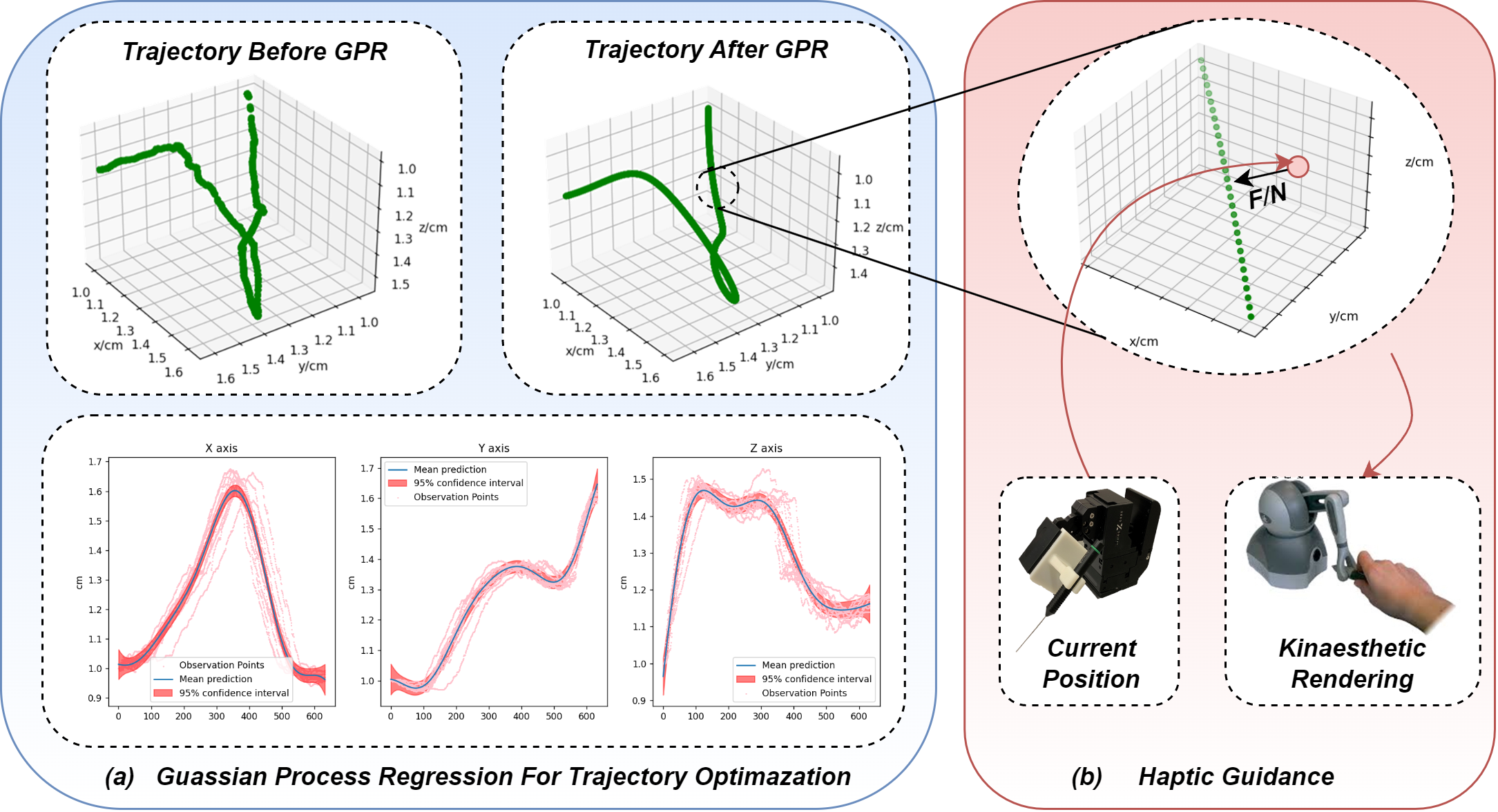}
	\caption{Illustration of the Haptic Guidance. (a) shows the GPR for Trajectory Optimization. (b) shows how haptic guidance force is generated.}
   \vspace{-0.5cm}
	\label{fig:Gpr}
\end{figure*}

Haptic guidance can be a useful tool in training individuals to manipulate robots along predetermined trajectories. In this study, we employed the learning from the demonstration paradigm to generate desired trajectories for surgical training \cite{chen2020supervised}.
Specifically, we collected manipulation trajectories from experienced operators and optimized them using Gaussian Process Regression (GPR). To prepare the trajectories for GPR training, we recorded ten trajectories as the training set with ROS, capturing the 3D coordinate ($\bm{p}=[x, y, z]$) of the microrobot's end effector.
We then preprocessed the trajectories by removing duplicate points and downsampling them to N points to ensure that all trajectories had the same length for GPR training. We utilized the GaussianProcessRegressor from scikit-learn \cite{scikit-learn} to regress those ten trajectories. We set the observation values as an integer sequence from 0 to $N$, and the predicted values as a series of corresponding coordinates $\bm{p_t}(t=0,1,...,N)$ for each data position along the trajectory.  The radial basis function (RBF) kernel was used for the GPR to regress a smooth trajectory. Fig. \ref{fig:Gpr} (a) shows the GPR for trajectory optimization. The resulting 95\% confidence interval was small (within 0.10 cm), and the trajectory was smooth, indicating the success of the optimization process.



Fig. \ref{fig:Gpr}(b) illustrates how the haptic guidance force is generated based on the robot's current position and predetermined trajectory.
Within the ROS framework, we subscribe to the 3D position of the microrobot's end effector. Upon receiving a position $\bm{p}$, we proceed to calculate the Euclidean distance from this position point to all points on the predetermined trajectory. Subsequently, we identify the coordinates of the point $\bm{d}$ on the trajectory with the minimum distance to $\bm{p}$.
If the minimum distance exceeds our predefined threshold ($0.2 mm$), force feedback is provided to the operator. This force vector is obtained by subtracting $\bm{d}$ from $\bm{p}$, as the vector formed by two position points represents the guidance force with direction and magnitude.
Finally, the resultant vector is transmitted to the Geomagic Touch, which offers continuous force feedback for real-time haptic guidance.

\section{User Studies}

\subsection{Experiment Design}
\begin{figure*}[htbp]
    \centering
    \captionsetup{font=footnotesize,labelsep=period}
    \includegraphics[width = 1\hsize]{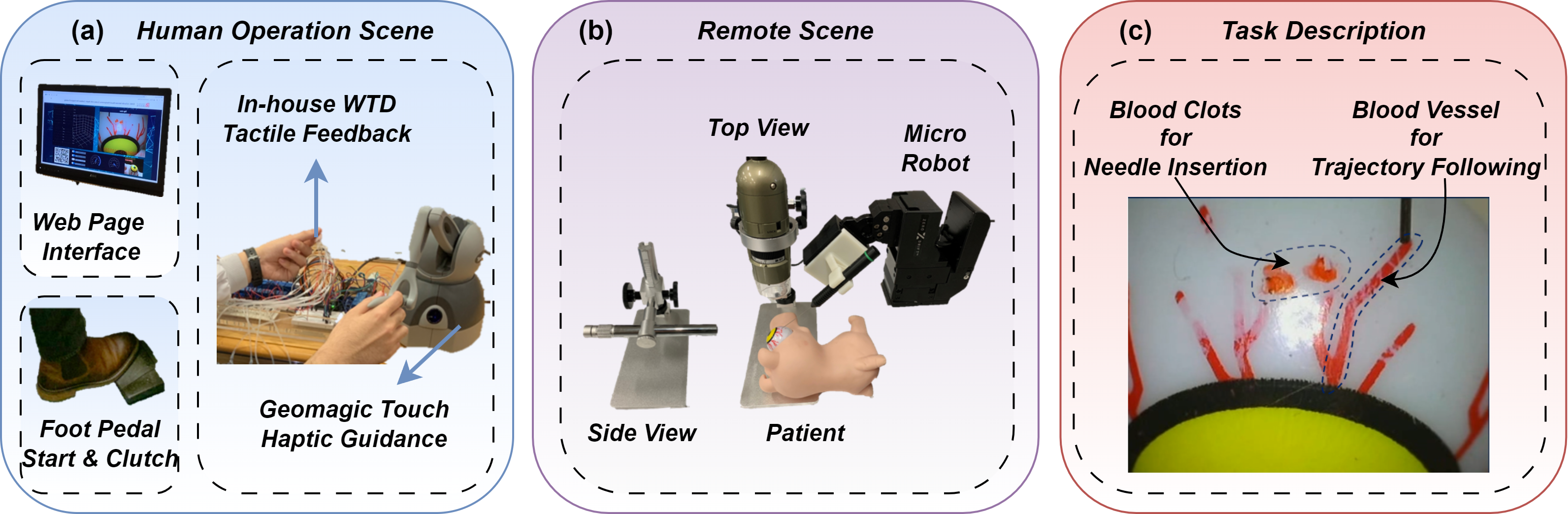}
    \caption{Overview of the experiment setup for user studies.  Overview of  (a) the human operation scene;  (b) the microsurgical robot platform in the remote scene. (c) visualization of the eyeball phantom and the microsurgical task description. }
      \vspace{-0.4cm}
    \label{fig:Setup}
\end{figure*}

In order to evaluate the effectiveness of our proposed TIMS, we conducted a user study with the experimental setup shown in Fig. \ref{fig:Setup}. A total of five participants (3 females and 2 males) were invited to join the study. Two participants have prior experience playing video games, while two are familiar with Geomagic Touch. 
We employed two micro-surgical tasks for user studies. Ethical approval was obtained from the University of Bristol Ethics Committee (ref No.10389) prior to conducting the study.


\textbf{Trajectory Following}: This task was designed to test the trainee's ability to perform general position-changing maneuvers. The trainee was required to trace a pre-defined trajectory, as shown in Fig. \ref{fig:Setup}(c). The trajectory represents a blood vessel in a simulated eyeball, where the trainees were required to navigate with high precision. They needed to avoid any excessive force on the eyeball while maintaining contact between the surgical tool and the blood vessel.



\textbf{Needle Insertion}: This task was designed to test the trainees' ability to locate a desired position for precise operation. The trainee was instructed to move the surgical tool to the appropriate position and use the `insertion' function on the stylus of Geomagic Touch to puncture two blood clots in the simulated eyeball, as shown in Fig. \ref{fig:Setup}(c). The task required the trainees to perform the insertion function with high accuracy while avoiding damage to the surrounding tissue.

To assess the effectiveness of our system, we designed four experiments for each participant. These include operations with i) No Feedback (NF), ii) Tactile Feedback only (TF), iii) Haptic Guidance only (HG), and iv) both Tactile Feedback and Haptic Guidance (TF \& HG). 
The microsurgical tool will be moved back to the original position after the trainees finish all the procedures of the user studies. This can ensure a fair comparison among different system setups.  

Prior to the formal experiment, each participant was allowed to familiarize themselves with the entire test procedure through a practice session. During the formal user studies, each participant was asked to perform two trials under the same operation condition, resulting in a total of eight trials for both the Trajectory Following and Needle Insertion tasks (4 different experiments x 2 trials). The ROS-Django system automatically recorded all operation data, including the operation trajectory, time, and foot pedal messages. These data can be used for performance evaluation and skill analysis. After each experiment, participants were asked to complete a questionnaire consisting of six questions with scores ranging from 1 to 5. Higher scores indicated greater satisfaction with the system and feedback function. The questions were as follows:




Q1: How challenging did you find it to control the system?

Q2: Was the system latency (live video and control) acceptable?

Q3: How difficult was it to complete the trajectory following task?

Q4: How difficult was it to determine whether a microsurgical tool had touched the eyeball or not?

Q5: When tactile feedback was applied, how effective was the information provided in indicating whether the tool touched the eyeball or not?

Q6: When haptic guidance was applied, how effective was the guidance in helping you to complete the operation?






To quantify the effectiveness of TIMS, we conducted comparative studies based on four evaluation metrics:
\begin{itemize}
    \item \textbf{Reminder Times:} If the system does not provide any feedback, participants may inadvertently apply excessive force through the microsurgical tool, causing potential damage to the eyeball. To mitigate this risk, we provided reminders to participants if necessary, and the number of reminders was recorded.
    \item \textbf{Time Cost:} For each experimental setup, we calculated the average time it took for participants to successfully complete two tasks: trajectory following and needle insertion. This metric reflects the difficulty of the operations under different experimental settings.
    \item \textbf{Trajectory Following Error:} The trajectory following error is calculated based on root mean square error (RMSE), which represents the square root of the discrepancy between the desired trajectory and the actual trajectory. 
     \item    \textbf{Insertion Error:} This metric is designed to evaluate participants' performance in the needle insertion task. Insertion error is determined by the Euclidean distance between the desired target and the actual insertion position generated by the microsurgical tool.
\end{itemize}

\subsection{Results Analysis}
\subsubsection{Qualitative Analysis}

Table \ref{tab:Questionnaire} shows the results of the questionnaire. When there was \textbf{No Feedback (NF)}, participants subjectively perceived the system to be challenging to control, with difficulty in following the desired trajectory and determining whether the surgical tool touches the eyeball or not. 
With \textbf{Tactile Feedback (TF)}, the difficulty of trajectory following and tool contact judgment has significantly reduced (Q3 score is 1.0 higher, and Q4 score is 2.85 higher on average compared to the No Feedback condition). Participants simply focused on the top view  without needing to split their attention to the depth perception of the microsurgical tool.
With \textbf{Haptic Guidance (HG)}, Q3 score is significantly improved compared to both the No Feedback and Tactile Feedback, while Q4 score has a slight improvement. During trajectory following tasks, Haptic Guidance generates force to pull the trainee's operation trajectory back onto the desired trajectory, which performs a similar function as TF. Unfortunately, the advantage of Haptic Guidance was not evident for the needle insertion task. It is still difficult for participants to determine whether the microsurgical tool is in contact with the eyeball or not. 
When the \textbf{Tactile Feedback and Haptic Guidance (TF\&HG)} were both activated, the highest score was achieved for all questions, indicating that the combination of the two feedback modes could provide participants with a better operating experience.  

Table \ref{tab:Questionnaire} indicates that participants found both Tactile Feedback and Haptic Guidance have a positive impact on microsurgical operations, with an average score of 4 out of 5. Additionally, Q2 scores suggest that participants found the delay in TIMS to be acceptable, with an average score of 4.04 across all experimental settings.





\begin{table}[htbp]
\centering
\captionsetup{font=footnotesize,labelsep=period}
\caption{Questionnaire Result}
\label{tab:Questionnaire}
\begin{tabular}{@{}lcccccc@{}}
\toprule
Task Settings & \multicolumn{1}{l}{Q1} & \multicolumn{1}{l}{Q2} & \multicolumn{1}{l}{Q3} & \multicolumn{1}{l}{Q4} & \multicolumn{1}{l}{Q5} & \multicolumn{1}{l}{Q6} \\ \midrule
No Feedback(NF)    & 2.14 & \multirow{4}{*}{4.04} & 1.71 & 1.29 & N/A  & N/A  \\
Tactile Feedback(TF)    & 3.29 &                       & 2.71 & 4.14 & 4.29  & N/A \\
Haptic Guidance(HG)    & 3.67 &                       & N/A & 2.33 & N/A & 4.50  \\
Introduce Both(TF\&HG) & \textbf{4.00} &                       & \textbf{4.43} & \textbf{4.14} & \textbf{4.33} & \textbf{4.50} \\ \bottomrule
\end{tabular}
\end{table}

\subsubsection{Quantitative Analysis}

\begin{figure}[h]
    \centering
    \captionsetup{font=footnotesize,labelsep=period}
    \includegraphics[width = 1\hsize]{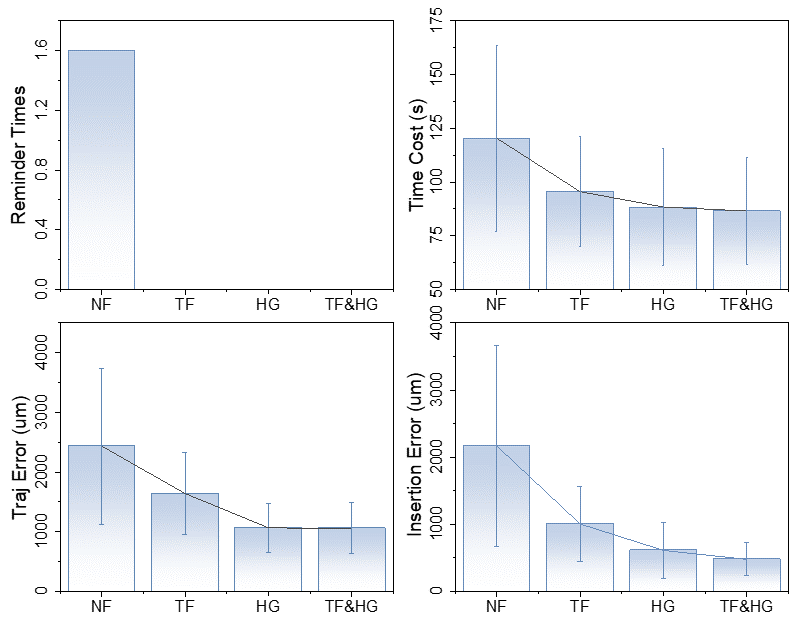}
    \caption{Visualization of the quantitative evaluation results using different experimental settings. `NF', `TF', `HG' represent `No Feedback', `Tactile Feedback', and `Haptic Guidance', respectively.}
      \vspace{-0.5cm}
    \label{fig:experiment result}
\end{figure}

The related metrics for trajectory following and needle insertion tasks completed by participants under different experimental settings are shown in TABLE \ref{tab:my-table} and Fig. \ref{fig:experiment result}. 
Poor performance was observed under the No Feedback condition. When Tactile Feedback or Haptic Guidance was introduced, participants' task performance improved significantly, with Haptic Guidance showing greater improvement than Tactile Feedback.

More specifically, compared to the No Feedback condition, the introduction of Tactile Feedback resulted in a 20.63\% reduction in time cost, a 32.63\% reduction in trajectory error, and a 51.94\% reduction in insertion error.  In contrast, the introduction of Haptic Guidance led to a 26.59\% decrease in time cost, a 56.20\% decrease in trajectory error, and a 69.59\% decrease in insertion error.  When both types of feedback were provided to the operators, the performance was similar to the ones with Haptic Guidance in terms of trajectory error and time cost (within 2\%), but better in terms of insertion error (a 24.47\% decrease). Moreover, except for No Feedback, the reminder times were zero for all other conditions.



\begin{table}[]
\centering
\captionsetup{font=footnotesize,labelsep=period}
\caption{Experiment Results (Average across Participants)}
\label{tab:my-table}
\begin{tabular}{@{}lllll@{}}
\toprule
\begin{tabular}[c]{@{}l@{}}Task\\ Settings\end{tabular} &
  \begin{tabular}[c]{@{}l@{}}Remind\\ Times\end{tabular} &
  \begin{tabular}[c]{@{}l@{}}Time\\ Cost(s)\end{tabular} &
  \begin{tabular}[c]{@{}l@{}}Trajectory\\ Error(um)\end{tabular} &
  \begin{tabular}[c]{@{}l@{}}Insertion\\ Error(um)\end{tabular} \\ \midrule
No Feedback(NF)        & 1.6 & 120.43          & 2437.02          & 2136.67         \\
Tactile Feedback(TF)   & 0   & 95.58          & 1641.77          & 1026.88         \\
Haptic Guidance(HG)    & 0   & 88.41          & 1067.43          & 649.71          \\
Introduce Both(TF\&HG) & 0   & \textbf{86.55} & \textbf{1057.45} & \textbf{490.73} \\ \bottomrule
\end{tabular}
 \vspace{-0.4cm}
\end{table}

\subsection{Microsurgical Skill Assessment and Feedback}
\begin{figure}[h]
    \centering
\captionsetup{font=footnotesize,labelsep=period}
    \includegraphics[width = 1\hsize]{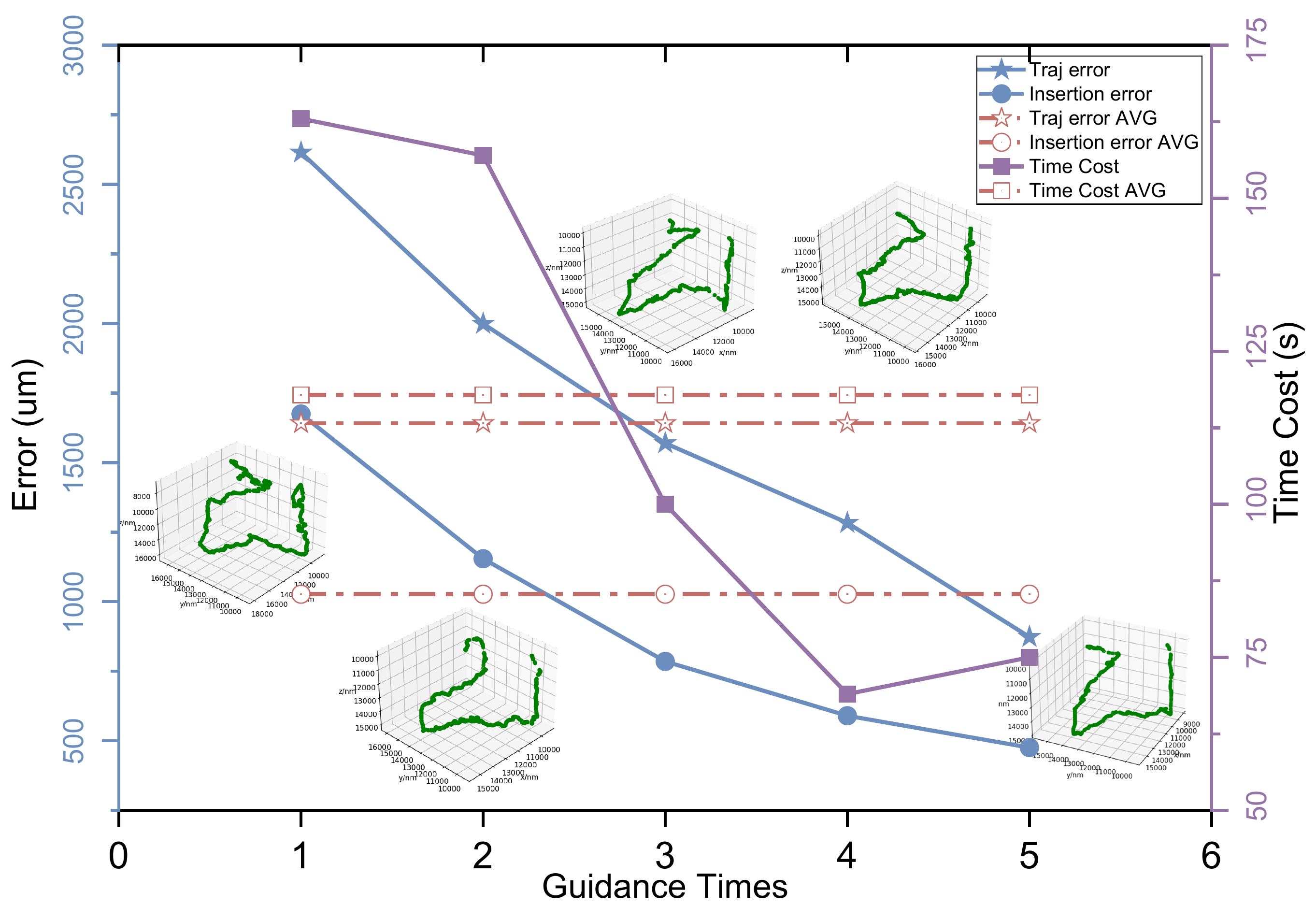}
    \caption{Visualization of the trainee learning curve. }
      \vspace{-0.4cm}
    \label{fig:assessment}
\end{figure}

We hypothesized that subjects would demonstrate improvement after using the proposed system with haptic guidance. We utilized standardized metrics to evaluate the effectiveness of haptic guidance for microsurgical skill training. We quantified microsurgical skill improvement after each task performed with haptic guidance. Specifically, operators were required to perform trajectory following and needle insertion ten times with tactile feedback.  We visualize the learning curve in Fig.  \ref{fig:assessment} for one of the participants as an example to demonstrate the value of haptic guidance in microsurgical training.   Following three rounds of haptic guidance, the trainee's performance exceeded the average performance quickly. The participant's performance continued to improve and stabilized after the fourth round of guidance.





\subsection{Discussions}
The user study results indicated that both tactile feedback and haptic guidance helped the operators improve performance during microsurgical training. Participants expressed great satisfaction with the TIMS system when both tactile feedback and haptic guidance were engaged. According to the quantitative analysis based on evaluation metrics, tactile feedback reduced the cognitive workload for participants significantly, as users did not need to repeatedly switch between top and side views to ensure that the distance between the tip of the microsurgical tool and the eyeball was within a safe distance. Haptic guidance enabled participants to follow the predetermined trajectory more precisely, thus reducing insertion and trajectory errors.




Regarding the metric of `reminder times', even though the web page displayed both the top view and side view of the microsurgery operation scene simultaneously, participants' attention remained on the top view.
 Hence, without haptic feedback, participants had higher risks of moving the surgical tool excessively in the z-axis direction, potentially injuring the patient's eyes during clinical operations. Therefore, we can conclude that both tactile feedback and haptic guidance are both significant for RAMS.

\section{Conclusions and Future Works}

In this paper, we develop a TIMS system that combines tactile internet and haptic guidance to enhance the efficiency of microsurgical training. We utilized deep learning methods to achieve vision-to-tactile information conversion, which eliminates the need for assembling expensive micro-force sensors onto the microsurgical tooltip, making the system more affordable for tiny clinical centres \cite{zhang2020hamlyn}. The touch perception for trainees is enabled by a WTD, which is an essential component of the tactile internet. We used ROS-Django to connect the leader (human operator) and the follower robot over the internet and built a web page interface to allow real-time visualization of the operating scenes. In addition, haptic guidance is integrated since it can provide operators with real-time force feedback based on the deviation between the current surgical tool position and the optimized trajectory, which can significantly enhance the efficiency of microsurgical training. Microsurgical skill analysis is provided by the system to help trainees monitor their progress after each training session.  User studies demonstrated the effectiveness of our proposed system and the significance of the tactile internet and 
haptic guidance function.

In the future, we aim to optimize the trajectory generation method by implementing advanced imitation learning to generate trajectories based on visual observations and incorporate trustworthiness features \cite{zhang2021explainable}. Furthermore, we plan to integrate Mixed Reality technology into the TIMS framework, which can provide trainees with an immersive operation experience. Additionally, more comprehensive evaluations will be provided to the users. For example, the `Structured
Assessment of Robotic Microsurgical Skills' scoring framework will be integrated into the training system \cite{alrasheed2014robotic}.
We will invite surgeons to participate in user studies and perform more complicated tasks such as microvascular anastomosis, membrane peeling, etc.



\bibliographystyle{IEEEtran}
\bibliography{references}

\begin{thebibliography}{10}
\providecommand{\url}[1]{#1}
\csname url@samestyle\endcsname
\providecommand{\newblock}{\relax}
\providecommand{\bibinfo}[2]{#2}
\providecommand{\BIBentrySTDinterwordspacing}{\spaceskip=0pt\relax}
\providecommand{\BIBentryALTinterwordstretchfactor}{4}
\providecommand{\BIBentryALTinterwordspacing}{\spaceskip=\fontdimen2\font plus
\BIBentryALTinterwordstretchfactor\fontdimen3\font minus
  \fontdimen4\font\relax}
\providecommand{\BIBforeignlanguage}[2]{{%
\expandafter\ifx\csname l@#1\endcsname\relax
\typeout{** WARNING: IEEEtran.bst: No hyphenation pattern has been}%
\typeout{** loaded for the language `#1'. Using the pattern for}%
\typeout{** the default language instead.}%
\else
\language=\csname l@#1\endcsname
\fi
#2}}
\providecommand{\BIBdecl}{\relax}
\BIBdecl

\bibitem{zhang2022teleoperation}
D.~Zhang, W.~Si, W.~Fan, Y.~Guan, and C.~Yang, ``From teleoperation to
  autonomous robot-assisted microsurgery: A survey,'' \emph{Machine
  Intelligence Research}, vol.~19, no.~4, pp. 288--306, 2022.

\bibitem{zhang2019handheld}
D.~Zhang, Y.~Guo, J.~Chen, J.~Liu, and G.-Z. Yang, ``A handheld master
  controller for robot-assisted microsurgery,'' in \emph{2019 IEEE/RSJ
  International Conference on Intelligent Robots and Systems (IROS)}.\hskip 1em
  plus 0.5em minus 0.4em\relax IEEE, 2019, pp. 394--400.

\bibitem{payne2021shared}
C.~J. Payne, K.~Vyas, D.~Bautista-Salinas, D.~Zhang, H.~J. Marcus, and G.-Z.
  Yang, ``Shared-control robots,'' \emph{Neurosurgical Robotics}, pp. 63--79,
  2021.

\bibitem{gao2021progress}
A.~Gao, R.~R. Murphy, W.~Chen, G.~Dagnino, P.~Fischer, M.~G. Gutierrez,
  D.~Kundrat, B.~J. Nelson, N.~Shamsudhin, H.~Su \emph{et~al.}, ``Progress in
  robotics for combating infectious diseases,'' \emph{Science Robotics},
  vol.~6, no.~52, p. eabf1462, 2021.

\bibitem{lallas2012robotic}
C.~D. Lallas, Davis, and J.~W. Members of the Society~of Urologic
  Robotic~Surgeons, ``Robotic surgery training with commercially available
  simulation systems in 2011: a current review and practice pattern survey from
  the society of urologic robotic surgeons,'' \emph{Journal of endourology},
  vol.~26, no.~3, pp. 283--293, 2012.

\bibitem{puliatti2020training}
S.~Puliatti, E.~Mazzone, and P.~Dell’Oglio, ``Training in robot-assisted
  surgery,'' \emph{Current Opinion in Urology}, vol.~30, no.~1, pp. 65--72,
  2020.

\bibitem{malik2017acquisition}
M.~M. Malik, N.~Hachach-Haram, M.~Tahir, M.~Al-Musabi, D.~Masud, and P.-N.
  Mohanna, ``Acquisition of basic microsurgery skills using home-based
  simulation training: A randomised control study,'' \emph{Journal of Plastic,
  Reconstructive \& Aesthetic Surgery}, vol.~70, no.~4, pp. 478--486, 2017.

\bibitem{choque2018virtual}
J.~Choque-Velasquez, R.~Colasanti, J.~Collan, R.~Kinnunen, B.~R. Jahromi, and
  J.~Hernesniemi, ``Virtual reality glasses and “eye-hands blind technique”
  for microsurgical training in neurosurgery,'' \emph{World neurosurgery}, vol.
  112, pp. 126--130, 2018.

\bibitem{ilie2008training}
V.~G. Ilie, V.~I. Ilie, C.~Dobreanu, N.~Ghetu, S.~Luchian, and D.~Pieptu,
  ``Training of microsurgical skills on nonliving models,'' \emph{Microsurgery:
  Official Journal of the International Microsurgical Society and the European
  Federation of Societies for Microsurgery}, vol.~28, no.~7, pp. 571--577,
  2008.

\bibitem{zhang2020microsurgical}
D.~Zhang, J.~Chen, W.~Li, D.~Bautista~Salinas, and G.-Z. Yang, ``A
  microsurgical robot research platform for robot-assisted microsurgery
  research and training,'' \emph{International journal of computer assisted
  radiology and surgery}, vol.~15, pp. 15--25, 2020.

\bibitem{zhang2022human}
D.~Zhang, Z.~Wu, J.~Chen, R.~Zhu, A.~Munawar, B.~Xiao, Y.~Guan, H.~Su, W.~Hong,
  Y.~Guo \emph{et~al.}, ``Human-robot shared control for surgical robot based
  on context-aware sim-to-real adaptation,'' in \emph{2022 International
  Conference on Robotics and Automation (ICRA)}.\hskip 1em plus 0.5em minus
  0.4em\relax IEEE, 2022, pp. 7694--7700.

\bibitem{giri2021application}
G.~S. Giri, Y.~Maddahi, and K.~Zareinia, ``An application-based review of
  haptics technology,'' \emph{Robotics}, vol.~10, no.~1, p.~29, 2021.

\bibitem{chen2020supervised}
J.~Chen, D.~Zhang, A.~Munawar, R.~Zhu, B.~Lo, G.~S. Fischer, and G.-Z. Yang,
  ``Supervised semi-autonomous control for surgical robot based on bayesian
  optimization,'' in \emph{2020 IEEE/RSJ International Conference on
  Intelligent Robots and Systems (IROS)}.\hskip 1em plus 0.5em minus
  0.4em\relax IEEE, 2020, pp. 2943--2949.

\bibitem{fettweis2014tactile}
G.~P. Fettweis, ``The tactile internet: Applications and challenges,''
  \emph{IEEE Vehicular Technology Magazine}, vol.~9, no.~1, pp. 64--70, 2014.

\bibitem{Piriyanont2013Design}
B.~Piriyanont, S.~R. Moheimani, and A.~Bazaei, ``Design and control of a mems
  micro-gripper with integrated electro-thermal force sensor,'' in \emph{2013
  Australian Control Conference}.\hskip 1em plus 0.5em minus 0.4em\relax IEEE,
  2013, pp. 479--484.

\bibitem{power2015cooperative}
M.~Power, H.~Rafii-Tari, C.~Bergeles, V.~Vitiello, and G.-Z. Yang, ``A
  cooperative control framework for haptic guidance of bimanual surgical tasks
  based on learning from demonstration,'' in \emph{IEEE International
  Conference on Robotics and Automation (ICRA)}.\hskip 1em plus 0.5em minus
  0.4em\relax IEEE, 2015, pp. 5330--5337.

\bibitem{chung2017affordable}
S.-B. Chung, J.~Ryu, Y.~Chung, S.~H. Lee, and S.~K. Choi, ``An affordable
  microsurgical training system for a beginning neurosurgeon: how to realize
  the self-training laboratory,'' \emph{World Neurosurgery}, vol. 105, pp.
  369--374, 2017.

\bibitem{zhang2019design}
D.~Zhang, J.~Liu, L.~Zhang, and G.-Z. Yang, ``Design and verification of a
  portable master manipulator based on an effective workspace analysis
  framework,'' in \emph{2019 IEEE/RSJ International Conference on Intelligent
  Robots and Systems (IROS)}.\hskip 1em plus 0.5em minus 0.4em\relax IEEE,
  2019, pp. 417--424.

\bibitem{zhang2020automatic}
D.~Zhang, Z.~Wu, J.~Chen, A.~Gao, X.~Chen, P.~Li, Z.~Wang, G.~Yang, B.~Lo, and
  G.-Z. Yang, ``Automatic microsurgical skill assessment based on cross-domain
  transfer learning,'' \emph{IEEE Robotics and Automation Letters}, vol.~5,
  no.~3, pp. 4148--4155, 2020.

\bibitem{zhang2018self}
D.~Zhang, B.~Xiao, B.~Huang, L.~Zhang, J.~Liu, and G.-Z. Yang, ``A
  self-adaptive motion scaling framework for surgical robot remote control,''
  \emph{IEEE Robotics and Automation Letters}, vol.~4, no.~2, pp. 359--366,
  2018.

\bibitem{wang2021real}
R.~Wang, D.~Zhang, Q.~Li, X.-Y. Zhou, and B.~Lo, ``Real-time surgical
  environment enhancement for robot-assisted minimally invasive surgery based
  on super-resolution,'' in \emph{2021 IEEE International Conference on
  Robotics and Automation (ICRA)}.\hskip 1em plus 0.5em minus 0.4em\relax IEEE,
  2021, pp. 3434--3440.

\bibitem{glenn_jocher_2021_5563715}
\BIBentryALTinterwordspacing
G.~J. et. al., ``{ultralytics/yolov5: v6.0 - YOLOv5n 'Nano' models, Roboflow
  integration, TensorFlow export, OpenCV DNN support},'' Oct. 2021. [Online].
  Available: \url{https://doi.org/10.5281/zenodo.5563715}
\BIBentrySTDinterwordspacing

\bibitem{scikit-learn}
F.~Pedregosa, G.~Varoquaux, A.~Gramfort, V.~Michel, B.~Thirion, O.~Grisel,
  M.~Blondel, P.~Prettenhofer, R.~Weiss, V.~Dubourg, J.~Vanderplas, A.~Passos,
  D.~Cournapeau, M.~Brucher, M.~Perrot, and E.~Duchesnay, ``Scikit-learn:
  Machine learning in {P}ython,'' \emph{Journal of Machine Learning Research},
  vol.~12, pp. 2825--2830, 2011.

\bibitem{zhang2020hamlyn}
D.~Zhang, J.~Liu, L.~Zhang, and G.-Z. Yang, ``Hamlyn crm: A compact master
  manipulator for surgical robot remote control,'' \emph{International journal
  of computer assisted radiology and surgery}, vol.~15, pp. 503--514, 2020.

\bibitem{zhang2021explainable}
D.~Zhang, Q.~Li, Y.~Zheng, L.~Wei, D.~Zhang, and Z.~Zhang, ``Explainable
  hierarchical imitation learning for robotic drink pouring,'' \emph{IEEE
  Transactions on Automation Science and Engineering}, 2021.

\bibitem{alrasheed2014robotic}
T.~Alrasheed, J.~Liu, M.~M. Hanasono, C.~E. Butler, and J.~C. Selber, ``Robotic
  microsurgery: validating an assessment tool and plotting the learning
  curve,'' \emph{Plastic and reconstructive surgery}, vol. 134, no.~4, pp.
  794--803, 2014.

\end{thebibliography}
\end{document}